# WaveletKernelNet: An Interpretable Deep Neural Network for Industrial Intelligent Diagnosis

Tianfu Li, *Student Member, IEEE*, Zhibin Zhao, Chuang Sun, Li Cheng,
Xuefeng Chen, *Senior Member, IEEE*, Ruqiang Yan, *Senior Member, IEEE*, Robert X. Gao, *Fellow, IEEE*

*Abstract*—Convolutional neural network (CNN), with ability of feature learning and nonlinear mapping, has demonstrated its effectiveness in prognostics and health management (PHM). However, explanation on the physical meaning of a CNN architecture has rarely been studied. In this paper, a novel wavelet driven deep neural network termed as WaveletKernelNet (WKN) is presented, where a continuous wavelet convolutional (CWConv) layer is designed to replace the first convolutional layer of the standard CNN. This enables the first CWConv layer to discover more meaningful filters. Furthermore, only the scale parameter and translation parameter are directly learned from raw data at this CWConv layer. This provides a very effective way to obtain a customized filter bank, specifically tuned for extracting defect-related impact component embedded in the vibration signal. In addition, three experimental verification using data from laboratory environment are carried out to verify effectiveness of the proposed method for mechanical fault diagnosis. The results show the importance of the designed CWConv layer and the output of CWConv layer is interpretable. Besides, it is found that WKN has fewer parameters, higher fault classification accuracy and faster convergence speed than standard CNN.

*Index Terms*—Prognostic and health management (PHM), convolutional neural network, continuous wavelet transform, continuous wavelet convolutional layer, machine fault diagnosis.

## I. Introduction

PROGNOSTIC and health management (PHM) in mechanical system has been applied widely to achieve predictive maintenance, including fault tracking, reduced downtime and asset preservation [1-3]. The rise of PHM has benefited from the development of advanced sensing technology, wireless communication and computing systems in recent years, which have generated a large amount of data for manufacturing systems [4], [5].

Traditionally, PHM consists of three main phases: sensor signal acquisition, feature extraction and selection, followed by fault classification and fault prediction. Sensor signals are collected from PHM system. Feature extraction and selection are the critical sections in fault diagnosis, even once became a research hotspot in PHM. For example, Li [6] proposed semi-supervised kernel Marginal Fisher analysis for feature extraction. It can catch sight of the inherent manifold structure of the data set, and consider the tightness inside the class and the separability between classes. Qiao [7] used empirical mode decomposition, fuzzy feature extraction and support vector machines to diagnose turbine generator sets under three different operating conditions. Chen [8] used the wavelet packet to extract features, and proposes an online monitoring model based on logistic regression to analyze the vibration signal of tool wear. In the last stage, machine learning models are trained by the extracted features and output the results of fault predictions. However, these traditional fault diagnosis methods have some limitations in PHM. First, traditional methods rely on manually selected features. If the extracted features are inadequate for the task, the performance of the final classification algorithm will be greatly degenerated. Second, the handcrafted features are time-consuming, especially when dealing with large amounts of data, and even if the model is trained using the extracted features, the efficiency of the final prediction or classification still needs to be improved.

Unlike the above methods, the Deep Learning (DL) approach provides an effective solution to overcome these limitations, in part because of their powerful capabilities in feature learning. The deep architecture has multiple hidden layers, this could allow the deep architecture learn the essential features hidden in the monitoring data and improve diagnostic accuracy. There are many kinds of neural network structures in deep learning, such as deep belief network (DBN) [9], [10], auto-encoder (AE) [11], [12] and convolutional neural network (CNN) [13], [14]. CNN as a kind of DL architecture has been broadly applied to many areas [16], such as natural language processing [17], speech recognition [18], computer vision [19].

In recent years, the DL methods using CNN have swept the field of fault diagnosis and produced many remarkable achievements [20-22]. From the perspective of network input, it can be classified into one-dimensional input CNN (1D-CNN) and two-dimensional input CNN (2D-CNN). For example, Eren [23] employed compact adaptive 1D-CNN classifier for a generic real-time induction bearing fault diagnosis system. Jia [24] proposed a novel 1D-CNN framework for the fault diagnosis of three imbalanced bearing datasets. Besides, Yang [25] implemented Fourier transform for the collected bearing signals and the obtained spectra were used as the input of the 1D-CNN for bearing fault diagnosis. For 2D-CNN, Ding [26] proposed a multi-scale feature mining method based on wavelet packet energy image and deep CNN for spindle bearing fault diagnosis.

Although these CNN based DL methods have achieved successful applications in PHM, there are still problems associated with them. A main limitation of CNNs in mechanical fault diagnosis is that they operate in a black box, which does not offer insight into how to make the final decision. This may not only challenges the credibility of the decision itself, but also

Tianfu Li, Zhibin Zhao, Chuang Sun, Li Cheng, Xuefeng Chen and Ruqiang Yan are with School of Mechanical Engineering, Xi'an Jiaotong University, Xi'an, Shaanxi 710049, China. Corresponding author: Ruqiang Yan, E-mail: yanruqiang@xjtu.edu.cn.

Robert X. Gao is with Department of Mechanical and Aerospace Engineering, Case Western Reserve University, Cleveland, OH 44106, USA.

limits further improvement of the CNN to accommodate a wider range of applications.

The exploration of the interpretability of CNN has not stopped. Alekseev [27] proposed an interpretable CNN call GaborNet. In this paper, Gabor wavelet is used as the convolution kernel of the neural network for image input, which makes the convolutional combination interpretability. Grezmak [28] proposed an interpretable CNN for gearbox fault diagnosis based on layer-wise relevance propagation. In order to find the physical meaning of convolutional (Conv) layer output, some studies have visualized the output of each layer of the CNN [29], [30]. Bojarski [31] proposed a novel CNN called PilotNet, and explain what it learns and how it makes its decisions by visualizing the feature map of each layer.

Nonetheless, all these interpretable neural networks are mostly for 2D input. In mechanical fault diagnosis, we often get 1D vibration signals, and there is no specific convolutional kernel design method to extract impulsive component, which is the essential signal component related to machine fault. The results of Ref. [32] show that the first convolutional layer is one of the most critical part of current waveform-based CNNs. The first layer of CNN not only handles data with high-dimensional, but is also more susceptible to the problem of vanishing gradient, especially when the network architecture is very deep. This will lead to the learned filters of CNN typically employ a noisy and uncoordinated multi-band shape, particularly in the case of few training samples. These learned filters certainly make sense for the neural network, but they cannot effectively represent the mechanical signals. Therefore, it is desired to design a network that can extract their features according to the characteristics of the signals.

A promising approach to overcome the difficulty is explored in this paper, where a continuous wavelet convolution layer (CWConv) layer is designed to replace the first convolution layer, which aims to help the CNNs discover filters with a certain physical meaning. Based on this, the wavelet driven CNN, termed as WaveletKernelNet (WKN), is proposed, which can make the output of the first layer of CNN interpretable. The filter bank characteristics of standard CNN depend on a set of randomly initialized parameters (each element of the filter vector can be directly learned). By comparison, WKN convolves the input signal with a cluster of parametrized wavelet dictionaries that accomplish wavelet transform. The scale and translation parameters are the only two parameters of the CWConv layer learned from input signal. This approach forces the first layer of the network to focus on extracting features of the input signals, making the result of network output has clear physical meaning and more robust to different data. The main contributions of our work can be summarized as follows:

1) WaveletKernelNet is designed to extract fault features from mechanical signals. The first convolution layer is replaced by the CWConv layer. This is equivalent to adding some constraints to the waveform of the filters. Therefore, the essential semantic information with a clear physical meaning can be extracted from high-dimensional data.
2) The designed CWConv layer can extract the characteristics of the mechanical signal by means of inner product matching. This makes the feature maps acquired in the first layer interpretable.
3) The proposed CWConv layer is generalizable, and it can be applied to any existing convolution-based network. Besides, the parameters in the first convolutional layer are drastically reduced, and the convergence speed of the neural network is accelerated.

The remainder of this paper is organized as: theoretical foundation about replacing the CWConv layer with standard one is introduced in section II, based on which a novel machine fault diagnosis pipeline is shown in Section III. In Section IV, three datasets were experimentally studied to verify the effectiveness of WKN. Besides, in contrast test its performance were compared with other methods. In Section V, we further discuss the interpretability and properties of WKN. Conclusions are shown in Section VI.

## II. THEORETICAL FOUNDATION

### A. CNN

CNN consists of an input layer, a number of hidden layers for feature extraction, and an output layer to output the results. Generally, the hidden layer contains a convolution (Conv) layer, a pooling layer, an activation layer, and a fully connected layer.

In the convolutional layer, each randomly initialized kernel convolves through the width and height of the input as is shown in Fig. 1. The *k*-th feature map of the *l*-th layer before the nonlinear activation has the feature value $h_k^l$, which can be denoted as:

$$h_k^l = w_k^l * x + b_k^l \tag{1}$$

where $w_k^l$ is the weight of *k*-th convolutional kernel of the *l*-th layer and $b_k^l$ denotes the bias. *x* is the input signal, * is the convolutional operator.

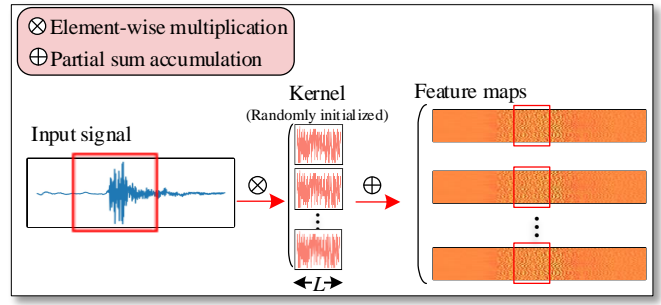

Fig. 1. Convolution process of the standard Conv layer in CNN.

An activation function is needed after the convolution operation. Since the feature map is nonlinearly activated, it can be denoted as:

$$z_k^l = f(h_k^l) \tag{2}$$

where $z_k^l$ is the activated feature maps.

After the convolutional layer, a large number of feature maps will be obtained, so the pooling layer is often used for down sampling operation to achieve the purpose of reducing the parameters of the model and retaining the main features. Mathematically, a pooling operation is defined as:

$$y_k^{l+1} = f(\beta^{l+1} down(z_k^l) + b^{l+1}) \quad (3)$$

where $down(\cdot)$ is the down sampling function, $\beta^{l+1}$ is the weight of the down sampling function of $(l+1\text{-}th)$ layer. If the average pooling with filter size is 2×2, the value of $\beta$ equals to 1/4.

After a series of stacked convolution and pooling layers, the final feature maps are flattened and then classified or regressed by the fully connected layer [33], [34]. The *softmax* function is often used for classification tasks to predict the probability of different categories. Cross entropy is often used as the loss function for CNN in model training, which can be denotes as:

$$H(r,p) = -\sum_i r_i log(p_i) \quad (4)$$

where $i$ is the number of class, $r \in \{0,1\}$ is the sample label, $p$ is the probability got from *softmax* function. The cross entropy loss is used to compute the error between the true label and the predicted probability. The gradient of the weight associated with the loss function is obtained by the back propagation algorithm, and the weight is updated by the stochastic gradient descent.

### B. Continuous Wavelet Transform

Convolution operation in CNN as shown in Eq. (1) can be expressed as the inner product of two vectors in signal processing. Also based on inner product, the principle of the continuous wavelet transform (CWT) is as follows.

In the temporal domain, a general wavelet dictionary $\{\psi_{u,s}(t)\}$ [35-37] can be defined by dilating the parameter $s > 0$, and translating the parameter $u \in R$ of the mother wavelet $\psi(\cdot)$ as follows:

$$\psi_{u,s}(t) = \frac{1}{\sqrt{s}} \psi(\frac{t-u}{s}) \quad (5)$$

where $t$ denotes the time, $s$ is a scale parameter that is inversely proportional to frequency, and $u$ is a translational parameter. The scale parameter $s$ can swell or compress the signal. When the scale parameter is relatively low, the signal shrinks more, resulting in a more detailed resulting graph. Besides, when the scale parameter is relatively high, the signal is stretched out which means that the resulting graph will be presented in less detail.

The CWT of signal $g = x(t)$ is obtained by calculating the inner product with $\psi^*_{s,u}(t)$ which can be expressed as [38]:

$$\text{CWT}_f(u,s) = \langle g, \psi_{u,s}(t) \rangle = \frac{1}{\sqrt{s}} \int x(t) \psi^*(\frac{t-u}{s}) dt \quad (6)$$

where $\psi^*(\cdot)$ denotes the complex conjugate of the mother wavelet $\psi(\cdot)$. Eq. (6) indicates that the CWT is similar to Fourier transform, where the basis function used by the Fourier transform to implement inner product matching is a trigonometric function. Besides, Fourier transform is full-band, so it does not have multi-scale properties, and the wavelet transform achieves multi-scale analysis of non-stationary signals by two parameters. The process of continuous wavelet transform is shown in Fig. 2, which contains three steps:

  a) Select a wavelet basis function, then using Eq. (6) to calculate the similarity coefficient with a portion of the signal;

  b) Shift the wavelet to the right by $u$ and continue to calculate the similarity coefficient until all parts of the signal are calculated;

  c) Change the scale parameter $s$ of the wavelet and repeat steps a) and b) until the analysis of all scales is completed.

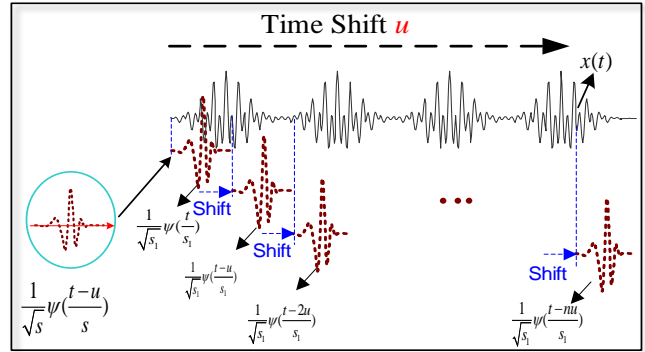

Fig. 2. Signal continuous wavelet transform processes.

After traversing all the scales, the signal $x(t)$ is decomposed into a series of similarity coefficients. At the same time, the wavelet dictionary is used to transform the signal $x(t)$ and project it to two-dimensional (2-D) time and scale dimensions.

### C. The Continuous Wavelet Convolutional Layer

The first layer of CNN is very important. The feature map it extracts often affects the entire model. Normally, the first layer of a standard CNN performs a set of time-domain convolutions between the input signal and some Finite Impulse Response (FIR) filters [39]. However, the impact components in the signal cannot be accurately extracted by convolution operations, besides, the extracted features also have no physical meaning and are not human readable.

To tackle this issue, according to the relationship between convolution operations and CWT, in WKN the first convolution layer is replaced by the CWConv layer, where the CWConv layer consist of wavelet kernels with different scale parameter and translation parameter as is shown in Fig. 3.

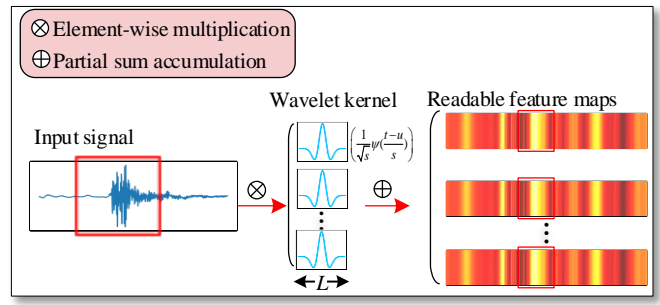

Fig. 3. The convolution process of the CWConv layer in CNN.

Conversely, the proposed WKN performs the convolution operation with a predefined function $\psi_{u,s}(t)$ that depends on two learnable parameters $\mu$ and $s$ only, as denoted in the following equation:

$$h = \psi_{u,s}(t) * x \quad (7)$$

where the predefined function $\psi_{u,s}(t)$ can be some wavelet functions with time-domain expressions, and here $h$ denotes the result of the CWConv operation.

In the back-propagation (BP) [40] process, the CWConv layer as the first layer only needs to update the parameters $u$, $s$, and we have:

$$\begin{cases} \delta_{u_k} = \dfrac{\partial H}{\partial u_k} = \dfrac{\partial H}{\partial z_k}\dfrac{\partial z_k}{\partial h_k}\dfrac{\partial h_k}{\partial \psi_{u,s}^k}\dfrac{\partial \psi_{u,s}^k}{\partial u_k} \\ \delta_{s_k} = \dfrac{\partial H}{\partial s_k} = \dfrac{\partial H}{\partial z_k}\dfrac{\partial z_k}{\partial h_k}\dfrac{\partial h_k}{\partial \psi_{u,s}^k}\dfrac{\partial \psi_{u,s}^k}{\partial s_k} \end{cases} \quad (8)$$

$$\begin{cases} u_k = u_k - \eta \delta_{u_k} \\ s_k = s_k - \eta \delta_{s_k} \end{cases} \quad (9)$$

where $\partial$ is the derivative operator, $\psi_{u,s}^k$ is the $k$-th wavelet filter of the first layer with length $L$, and $u_k, s_k$ represent its translation parameter and scale parameter, respectively. The parameter $u$, $s$ will be updated by subtracting the product of the learning rate $\eta$ and the gradient $\delta$.

For example, the partial derivative $\dfrac{\partial \psi_{u,s}}{\partial u}$ and $\dfrac{\partial \psi_{u,s}}{\partial s}$ of Morlet wavelet dictionary [42] can be obtained by the following steps.

First, the real part of Morlet wavelet is expressed as

$$\psi(t) = Ce^{-\frac{t^2}{2}}\cos(5t) \quad (10)$$

where $C$ is the normalized constant of signal reconstruction.

Second, a wavelet dictionary is obtained as

$$\psi_{u,s}(t) = \frac{1}{\sqrt{s}}Ce^{-\frac{(t-u)^2}{2s^2}}\cos(5\frac{t-u}{s}) \quad (11)$$

Third, according to the chain derivative rule, partial derivatives of parameter $s$ and parameter $u$ can be obtained, as shown below.

$$\begin{aligned} \frac{\partial \psi_{u,s}}{\partial u} &= \frac{C}{\sqrt{s}}e^{-\frac{(t-u)^2}{2s^2}}\frac{t-u}{s^2}\cos(5\frac{t-u}{s}) \\ &+ \frac{C}{\sqrt{s}}e^{-\frac{(t-u)^2}{2s^2}}\sin(5\frac{t-u}{s})\frac{5}{s} \end{aligned} \quad (12)$$

$$\begin{aligned} \frac{\partial \psi_{u,s}}{\partial s} &= -\frac{C}{2\sqrt[3]{s}}e^{-\frac{(t-u)^2}{2s^2}}\cos(5\frac{t-u}{s}) \\ &+ \frac{C}{\sqrt{s}}e^{-\frac{(t-u)^2}{2s^2}}\frac{(t-u)^2}{s^3}\cos(5\frac{t-u}{s}) \\ &+ \frac{C}{\sqrt{s}}e^{-\frac{(t-u)^2}{2s^2}}\sin(5\frac{t-u}{s})\frac{5(t-u)}{s^2} \end{aligned} \quad (13)$$

Therefore, the updates of parameter $s$ and parameter $u$ of Morlet wavelet convolutional layer can be made by bringing $\dfrac{\partial \psi_{u,s}}{\partial u}$ and $\dfrac{\partial \psi_{u,s}}{\partial s}$ into Eq. (8) and Eq. (9).

### III. MACHINE FAULT DIAGNOSIS USING WAVELETKERNELNET

A machine fault diagnosis pipeline is proposed in this section. The proposed pipeline can automatically learn the fault features and identify the working state of the machine from the original vibration signals. The overall procedure of the proposed mechanical fault diagnosis pipeline is shown in Fig. 4.

After vibration signals are acquired by sensors installed on machines during operation, they are truncated through a sliding window to form a set of sub-time series, which are used as the input of WKN. As the main purpose of this paper is to explore the value and significance of the CWConv layer, a relatively shallow backbone in Ref. [13] is used, where its first layer is replaced by CWConv. Finally, the WKN is trained by the training dataset, and the testing dataset is utilized to verify the effectiveness of WKN for fault diagnosis task.

Algorithm by applying WKN for mechanical fault diagnosis is summarized below.

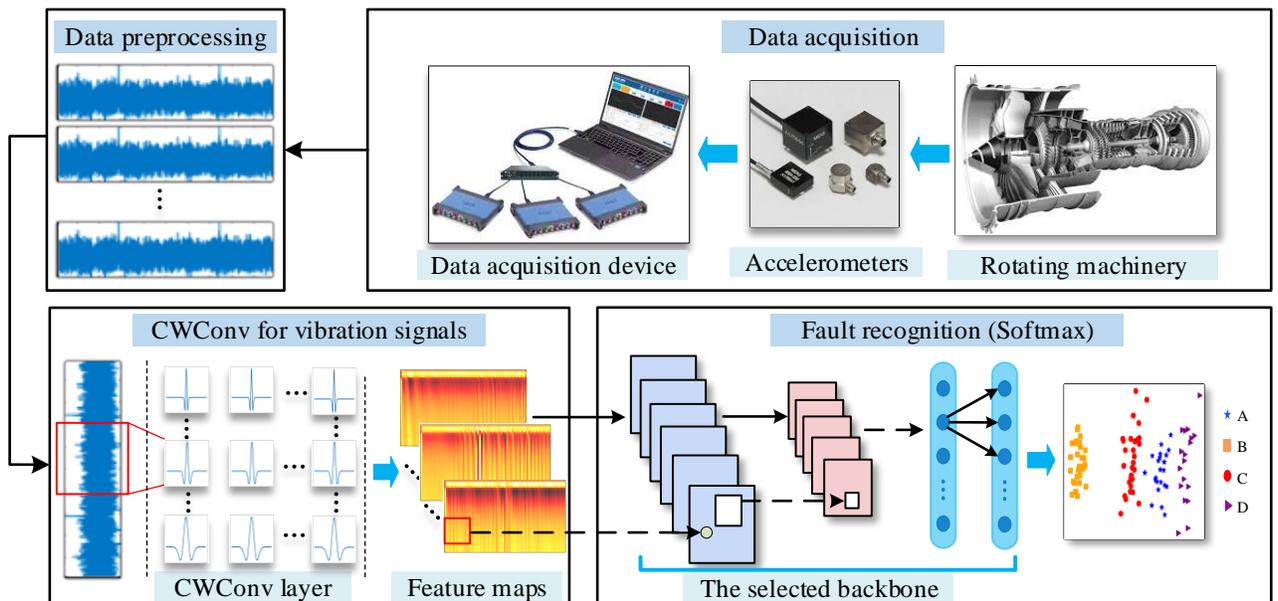

Fig. 4. General pipeline of mechanical fault diagnosis.

**Algorithm**: WaveletKernelNet for mechanical fault diagnosis

1. **Data preprocessing:** Maximum-Minimum normalization
2. **Model preparation**
   1) Select a CNN model as the backbone;
   2) Replace the first standard Conv layer by CWConv layer.
3. **Model training:**
   Input: Collected vibration signal dataset
   $X = [x_1, x_2, \cdots, x_n]$ with fault label $y_1, y_2, \cdots, y_n$
   1) Train WaveletKernelNet with the dataset $X$;
   2) Use Eq. (9) to update the first layer of the model, and use BP algorithm to update other layers of the model.
   Output: the WaveletKernelNet.
4. **Model validation:**
   Input: Validate set of the machine.
   Output: Fault category of the machine and classification accuracy.

## IV. EXPERIMENTAL VERIFICATION

To discuss the properties of WKN and validate its effectiveness, experiments on three datasets including bearing fault dataset, helical gear fault dataset and aeroengine bevel gear fault dataset were conducted. In addition, contrast experiments are also implemented to compare the performance of the proposed method with the standard CNN.

### A. Bearing Fault Dataset

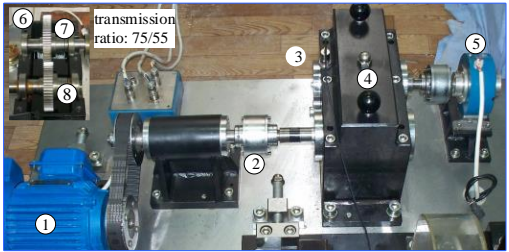

① drivemotor  ② shaft  ③ accelerometer  ④ gearbox  ⑤ load
⑥ inner structure of the gearbox  ⑦ fault/driven gear  ⑧ normal/driving gear

Fig. 5. Experimental facility.

The bearing fault dataset are acquired from the bearing fault simulation machine, shown in Fig. 5. The test machine mainly consist of the test machine drive motor, transmission part, gear fault simulation system, magnetic powder actuator, loading system, bearing fault simulation and dynamic balance simulation. The motor speed is controlled through the controller. This experiment mainly simulates the fault part of the bearing fault of the test bench.

The test was performed on 3,0205 tapered roller bearing with six prefabricated fault locations, including cage, rolling element, upper edge of outer ring, middle of outer ring, lower edge of outer ring, and inner ring. Fig. 6 illustrates some of the prefabricated faults. Together with the fault-free bearings, there are seven types of labels in total. In the experiment, the spindle speed was 1,200 r/min, and for each fault type, 2 minutes of vibration data with sampling rate of 96 kHz was collected.

For data preparation, a sliding window is used to truncate the vibration signal without overlap and each data sample contains 1,000 data points. After data preparation, the entire dataset is splited into training dataset and testing dataset, randomly. For each failure mode, it contains 7,474 data samples for model training and 1,869 data samples for model testing. Therefore, the training set have 52,318 data samples and the testing set have 13,083 data sample.

In this 7-class classification task, three kind of CWConv layer are designed as the first layer of WKN, they are formed by Mexhat wavelet [41], Morlet wavelet [42] and Laplace wavelet [43], respectively. In comparative experiments, we added a comparison experiment using the sin function as the first layer filter. In model settings, the length of the wavelet filter sets to 16 and use Adam as the optimizer. To verify the network stability and reduce the effects of the randomness, we take the average of 20 experimental results as the final result. The experimental results are shown in Table I and Fig. 7.

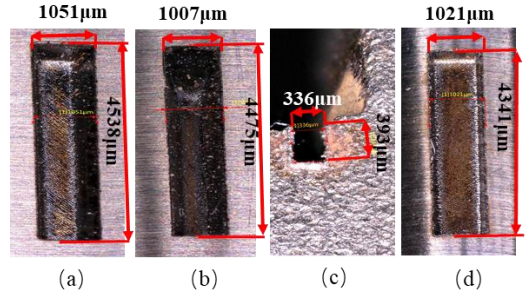

Fig. 6. Some fault types of tapered roller bearings. (a) outer ring fault; (b) inner ring fault; (c) cage fault; (d)rolling element fault.

As can be seen from these results, all the three types of WaveletKernelNets are able to achieve a higher classification accuracy and a smaller training loss than the standard CNN. Besides, we can see the LaplaceWaveletNet achieves the highest mean classification accuracy and has the fastest convergence speed among the three WaveletKernelNets. However, the SinCNN gets the lowest classification accuracy and the slowest convergence speed.

TABLE I.
CLASSIFICATION RESULTS OF BEARING DATASET

| Model | Mean accuracy | Variance |
| --- | --- | --- |
| MexhatWaveletNet | 92.61% | 0 |
| MorletWaveletNet | 92.85% | 0 |
| LaplaceWaveletNet | **99.91%** | **0** |
| CNN | 89.59% | 0.041 |
| SinCNN | 86.72% | 0.001 |

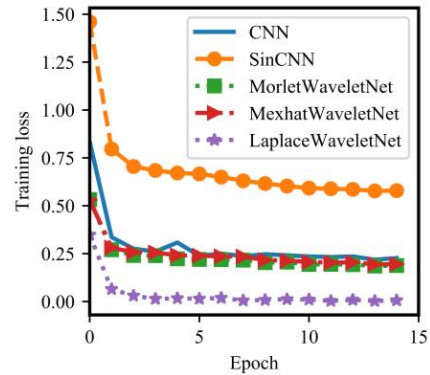

Fig. 7. Training loss on bearing dataset.

## B. Helical Gear Fault Dataset

The helical gear fault dataset was gathered from the gear fault simulation test rig as shown in Fig. 5. Vibration signals were collected while the rotating speeds increased from 1,000 to 1,040 rpm. During this experiment, it mainly tested two kinds of faults on the helical gear of the input shaft, including tooth surface wear and tooth root crack. Among them, tooth surface wear contains two failure modes and tooth root crack contains four failure modes. Together with the normal state, there are totally seven failure modes, as shown in Fig. 8.

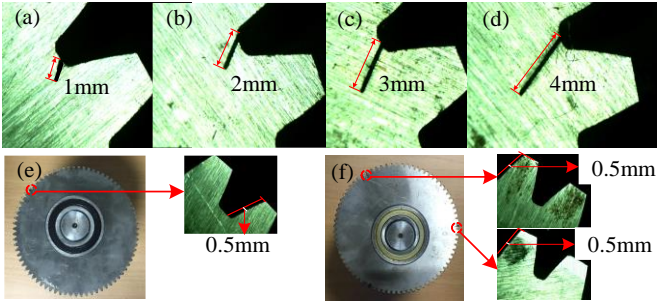

Fig. 8. Helical gear failure mode. (a)~(d) are tooth root crack. (e)~(f) are tooth surface wear.

Therefore, the helical gear fault diagnosis is considered as a 7-class classification task. The experiment settings are the same with the aforementioned bearing fault experiment. For each failure mode, it has 964 data samples for training, so for this 7-class classification task, the total number of training samples are 6,748 and the testing dataset contains 1,687 data samples.

The experimental results are shown in Table II and Fig. 9. As can be observed in those results, the mean classification accuracy of the three kind of WKNs are higher than traditional CNN and SinCNN. Besides, we can see that the convergence speed of SinCNN is still the slowest one.

TABLE II.
CLASSIFICATION RESULTS OF HELICAL GEAR DATASET

| Model | Mean accuracy | Variance |
|---|---|---|
| MexhatWaveletNet | 94.72% | 0 |
| MorletWaveletNet | 95.15% | 0 |
| LaplaceWaveletNet | **97.59%** | **0** |
| CNN | 83.45% | 0.081 |
| SinCNN | 77.21% | 0.002 |

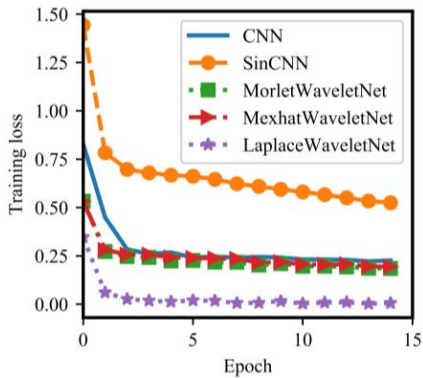

Fig. 9. Training loss on helical gear dataset.

## C. Aeroengine Bevel Gear Fault Dataset

Bevel gears shown in Fig. 10(a), as one of the transmission components of aero-engines, play a critical role in the reliability of aero-engine operations. In order to identify its failure mode, we collected data of four failure modes through the device shown in the Fig. 10(b) and Fig. 10(c). During the engine test, vibration signals of bevel gear were acquired by fixing acceleration sensors on the accessory casing corresponding to the bevel gear. Three different working conditions with rotating speed increased from 500rpm to 3,900rpm are investigated .

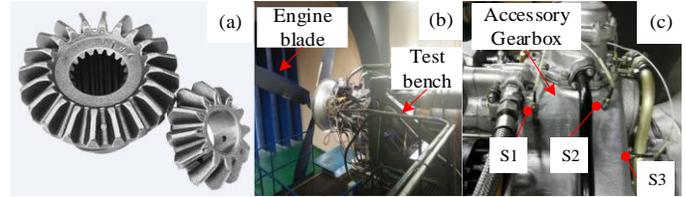

Fig. 10. Schematic diagram of the vibration test position of a certain aircraft engine accessory. (a) Aeroengine bevel gear. (b) A turboprop engine and, (c) sensor position layout.

For each working condition, the dataset contains three failure types and one health state, that is, broken teeth (BT), tooth surface wear (TSW), overhaul after returning to the factory (OAF), no fault (NF). Each fault mode under the same conditions has 800 data samples for model training and 200 data samples for model testing, thus there are 3,200 samples in model training and 800 samples for test in each experiment.

The experimental results are shown in Table III. The PCA visualization of classification results of LaplaceWaveletNet under four working conditions is shown in Fig. 11.

As can be seen in those results, in each working condition our WKN outperforms the traditional CNN and SinCNN. In addition, it can be found that the diagnostic accuracy of LaplaceWaveletNet is the highest one, which in turn indicates the LaplaceWaveletNet matches vibration components of the aeroengine bevel gear very well.

TABLE III.
CLASSIFICATION RESULTS OF AEROENGINE BEVEL GAR FAULT DATASET

| Working Condition | Model | Mean accuracy | Variance |
|---|---|---|---|
| 500rpm | MorletWaveletNet | 87.48% | 0 |
| | MexhatWaveletNet | 87.86% | 0 |
| | LaplaceWaveletNet | **98.85%** | **0** |
| | CNN | 80.58% | 0.009 |
| | SinCNN | 86.41% | 0 |
| 2000rpm | MorletWaveletNet | 97.98% | 0.001 |
| | MexhatWaveletNet | 95.20% | 0.006 |
| | LaplaceWaveletNet | **98.65%** | **0.001** |
| | CNN | 94.58% | 0.018 |
| | SinCNN | 88.92% | 0.005 |
| 3900rpm | MorletWaveletNet | 94.89% | 0 |
| | MexhatWaveletNet | 96.25% | 0 |
| | LaplaceWaveletNet | **99.42%** | **0** |
| | CNN | 91.97% | 0.010 |
| | SinCNN | 81.49% | 0.004 |

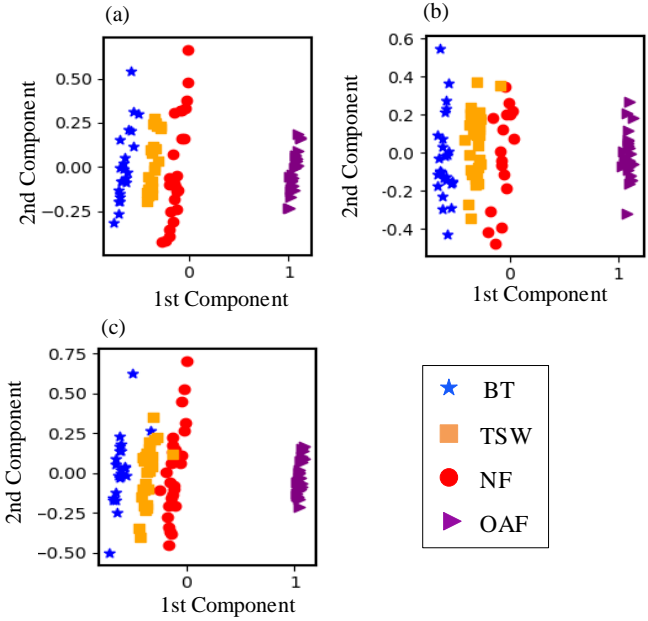

Fig. 11. PCA visualization of fault classification results in LaplaceWaveletNet under three working conditions. (a) working condition:500rpm; (b) working condition: 2,000rpm; (c) working condition: 3,900rpm.

## V. MODEL INTERPRETABILITY

In this section, we will discuss the interpretability and the convergence speed of the WKN. The interpretability of WKN will be illustrated from two aspects, that is, visualization of the feature map of CWConv layer and its waveforms, and the perspective of inner product matching.

### A. The Feature Map of CWConv Layer and Its Waveform

For a clear explanation of the CWConv layer, we use one data sample shown in Fig.12 as the input signal and discuss the difference between the feature map of three kinds of CWConv layer, traditional Conv layer and sin function filter. Moreover, the difference from their waveforms before training and after training is illustrated afterwards.

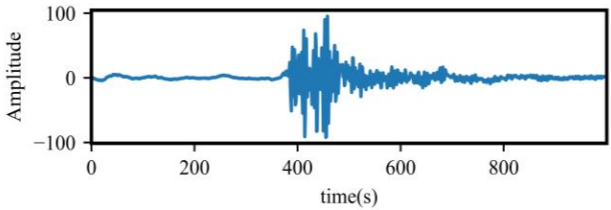

Fig. 12. Signal sample.

In this experiment, the networks were first trained by the bearing fault dataset. Then, the signal sample is input to this trained network, and extract the feature maps of the target layer. The extracted feature maps and its kernel waveforms are shown in Fig. 13 and Fig. 14, respectively.

For simplicity, the feature map of MexhatConv layer, MorletConv layer, LaplaceConv layer, traditional Conv layer and sin function layer are denoted as MexFM, MorFM, LapFM, TCFM, and SFM, respectively.

As can be seen from Fig. 13, the energy of MexFM and MorFM are dispersed in each frequency band of the signal. On the contrary, the LapFM, the TCFM and the SFM are more concentrated in the impact portion of the signal. However, the position and energy of the impact can be more clearly indicated on the LapFM. This can explain why LaplaceWaveletNet is able to perform the best in the three datasets.

Moreover, as can be seen in Fig. 14, the traditional CNN does not always learn filters with a well-defined waveform, and the waveform of the standard CNN looks noisy in some cases (see Fig. 14(g) and Fig. 14(h)). WKN, instead, has a fixed waveform that allows the resulting filter to have multi-scale and multi-resolution analysis capabilities through continuously learning the scale parameter and time-shift parameter from raw data. Therefore, WKN can match the input waveform very well and extract the characteristics of the signal effectively.

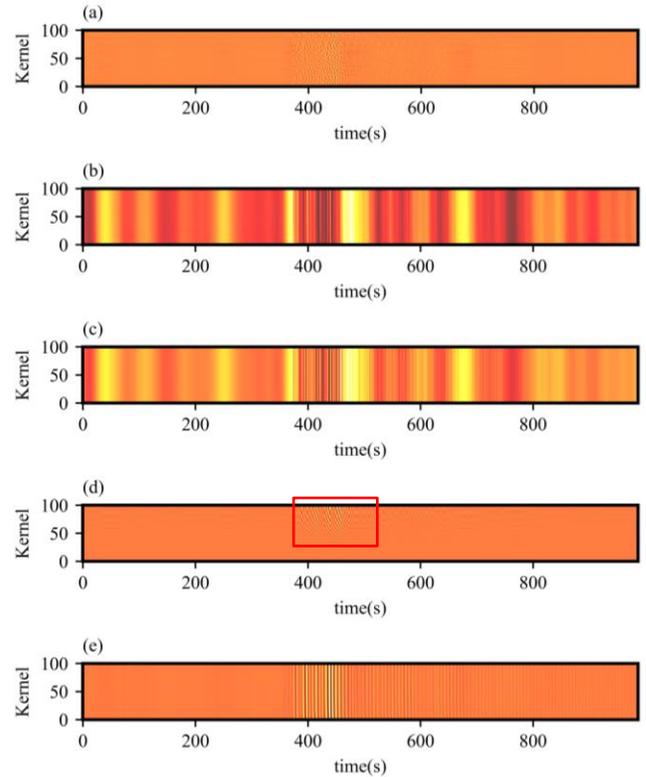

Fig. 13. The feature map of traditional CNN, WaveletKernelNet and SinCNN. (a) The feature map of conventional Conv layer; (b) The feature map of MorletWavelet layer; (c) The first feature map of MexhatWavelet layer; (d) The feature map of LaplaceWavelet layer; (e) The feature map of Sin function layer.

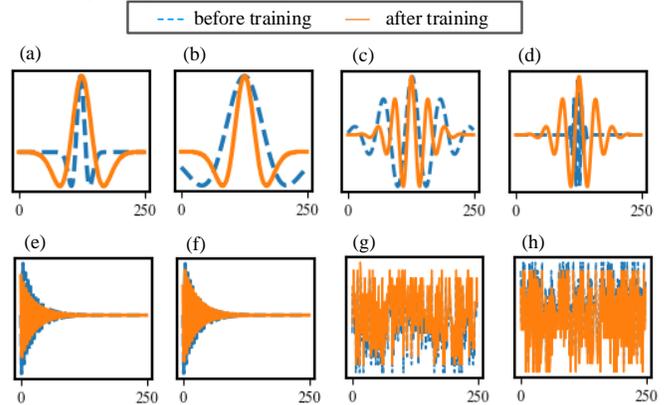

Fig. 14. Waveform of WaveletKernelNet and CNN before and after training. (a), (b) are the first and last waveforms of MexhhatWaveletNet. (c), (d) are the first and last waveforms of MorletWaveletNet. (e), (f) are the first and last waveforms of LaplaceWavelet; (g), (h) are the first and last waveforms of CNN.

*B. Inner Product Matching*

It can be observed from the previous experiment results that LapalceWaveletNet performs the best in the all those three datasets, while the SinCNN perform the worst. This can be explained from the perspective of inner product matching.

The main fault feature of the bearing and gear vibration signals are the periodic impulses. The Laplace wavelet is a kind of unilaterally attenuated wavelet which is mainly used to extract the features of the impact signal. Therefore, the designed LaplaceWaveletNet can match the impact component of the signal well, thus it can make the first layer of WKN extract more informative high-dimensional features. However, sin function is mainly used to extract the characteristics of stationary periodic signals, thus SinCNN can't match the impact component of the signal well. In deep neural network, this inappropriate filter, not only does not extract the valuable features, but also brings the wrong information, which makes the SinCNN perform even worse than the standard CNN.

*C. Hyperparameter and The Convergence Speed*

In the conventional convolutional layer, the number of parameters to be learned are the product of the number of filters $F$ and the length $L$ of the filter, that is, $F$ x $L$. However, a CWConv layer would only require $F$ x 2 parameters (a scale parameter, and a translation parameter are sufficient for design of each filter). Therefore, the parameters to be learned in the CWConv layer are reduced compared with the conventional convolutional layer, thus greatly accelerating the convergence speed of the network, the architecture comparison of CNN and WKN is shown in Table IV.

TABLE IV.
ARCHITECTURE COMPARISON BETWEEN CNN AND WAVELETKERNELNET

| CNN | | | WKN | | |
|---|---|---|---|---|---|
| Layer | Output | Param | Layer | Output | Param |
| Conv_1d | 1x100x985 | 1,700 | **CWConv** | **1x100x985** | **200** |
| Conv_1d | 1x6x981 | 3,006 | Conv_1d | 1x6x981 | 3,006 |
| Conv_1d | 1x16x486 | 496 | Conv_1d | 1x16x486 | 496 |
| Avg_pool | 1x16x16 | 0 | Avg_pool | 1x16x16 | 0 |
| Linear-1 | 1x120 | 30,840 | Linear-1 | 1x120 | 30,840 |
| Linear-2 | 1x84 | 10,164 | Linear-2 | 1x84 | 10,164 |
| Linear-3 | 1x7 | 595 | Linear-3 | 1x7 | 595 |

## VI. CONCLUSION

In this paper, a continuous wavelet convolution layer is presented to replace the first conventional convolution layer of the standard CNN to provide a solution to discover filters with certain physical meaning. Application of the presented WKN structure to machine fault diagnosis demonstrates its effectiveness. Conclusions of this paper is summarized as: 1) Using the physically meaningful wavelet convolutional layer as the first layer of the CNN improves the model interpretability; 2) Accuracy of machine fault diagnosis is enhanced compared with the standard CNN models, which indicates CWConv layer is able to improve performance of DL models. The experiment results show the CWConv layer is interpretable and has fewer parameters than standard Conv layer, which makes the WKN have faster convergence speed.